\pdfoutput=1

\documentclass[11pt]{article}

\usepackage{emnlp2021}

\usepackage{times}
\usepackage{latexsym}

\usepackage[T1]{fontenc}

\usepackage[utf8]{inputenc}

\usepackage{microtype}

\usepackage{url}
\usepackage{numprint}
\usepackage{amsmath}
\usepackage{multirow}
\usepackage[normalem]{ulem}
\usepackage{algpseudocode,algorithm,algorithmicx}
\DeclareRobustCommand{\bgg}[1]{{\sethlcolor{green}\hl{\textit{#1}}}}
\DeclareRobustCommand{\bgr}[1]{{\sethlcolor{red}\hl{\textit{#1}}}}

\usepackage{graphics,graphicx}
\usepackage{xcolor,soul}

\usepackage[colorinlistoftodos,prependcaption,textsize=tiny]{todonotes}
\usepackage{xargs}
\newcommandx{\gk}[2][1=]{\todo[backgroundcolor=orange!25,#1]{GK: #2}}
\newcommandx{\mg}[2][1=]{\todo[backgroundcolor=teal!25,#1]{GK: #2}}
\newcommandx{\he}[2][1=]{\todo[backgroundcolor=brown!25,#1]{GK: #2}}
\newcommandx{\jr}[2][1=]{\todo[backgroundcolor=blue!25,#1]{GK: #2}}

\title{Unsupervised and Distributional Detection of Machine-Generated Text}

\author{Matthias Gall\'e \and  Jos Rozen \and Germ\'an Kruszewski \and Hady Elsahar \\ Naver Labs Europe}

\begin{document}
\maketitle
\begin{abstract}
    The power of natural language generation models has provoked a flurry of interest in automatic methods to detect if a piece of text is human or machine-authored.
    The problem so far has been framed in a standard supervised way and consists in training a classifier on annotated data to predict the origin of one given new document.
    In this paper, we frame the problem in an \textit{unsupervised} and \textit{distributional} way: we assume that we have access to a large collection of unannotated documents, a big fraction of which is machine-generated.
    
    We propose a method to detect those machine-generated documents leveraging repeated higher-order $n$-grams, which we show over-appear in machine-generated text as compared to human ones.
    That weak signal is the starting point of a self-training setting where pseudo-labelled documents are used to train an ensemble of classifiers.
    
    Our experiments show that leveraging that signal allows us to rank suspicious documents accurately.
    Precision at \numprint{5000} is over $90\%$ for \texttt{top-k} sampling strategies, and over $80\%$ for \texttt{nucleus} sampling for the largest model we used (GPT2-large).
    The drop with increased size of model is small, which could indicate that the results hold for other current and future large language models.
\end{abstract}
\section{Introduction}

\begin{figure}
    \centering
    \includegraphics[width=.48\textwidth]{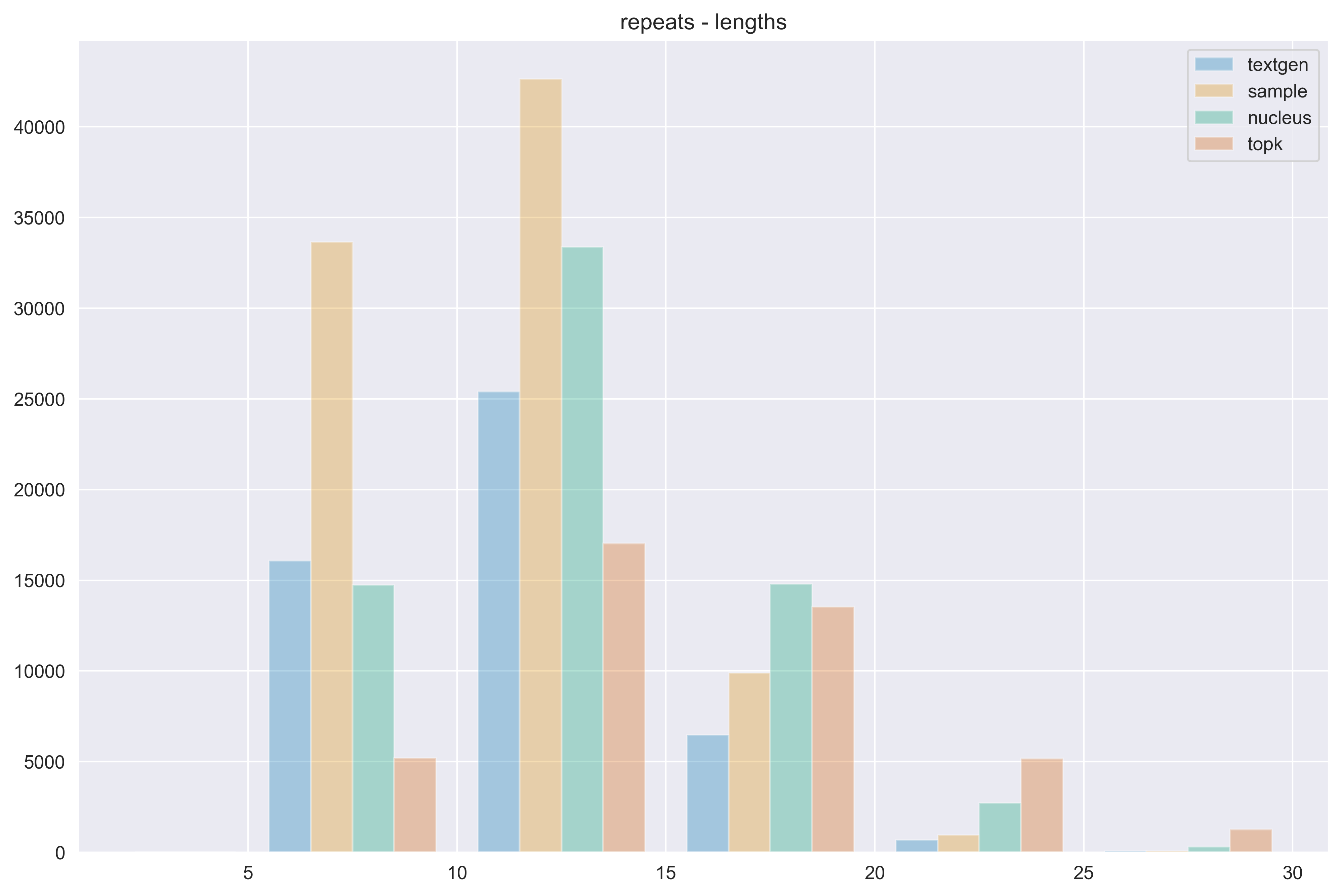}
    \caption{Histograms of the lengths of super-maximal repeats occurring in each dataset. \texttt{textgen} refers to human generated text.}
    \label{fig:repeatstats}
\end{figure}

The introduction of powerful neural left-to-right language models in the last few years has revolutionzed the capacity to automatically generate human-like text.
Their power has grown proportionally to the worry that those models could be used to generate \textit{fake} documents,\footnote{referring here to machine-generated text as opposed to text containing objectively false statements} either news articles~\citep{grover}, comments~\citep{readattendcomment} or other type of text~\citep{buchanan2021truth} without revealing the non-human authorship of it.
Recently, a number of papers have investigated the capacity of humans and learning algorithms to determine if a given text is human or machine generated~\citep{ippolito2020,dugan2020,maronikolakis2020}.
However, so far all those methods are making two key assumptions: (i) the settings is framed in a supervised way for the machine learning methods, having access at training time to machine-generated samples from the same generator; (ii) the decision is done for each sample individually and independently.

\smallskip

In this paper, we argue that this is not realistic. 

On one side any ill-intented adversary would take care of either retraining its own model, or modifying the generation in such a way as to drift significantly the distribution of generated text.
This can easily be done by using different sampling strategies~\citep{f2softmax,Welleck2020Neural,nucleus,massarelli-etal-2020-decoding}, an a-posterior re-weighting strategy or the use of one of a number of controlled generation methods currently available~\citep{plugplay,khalifa2020}. 
More-over, the domain used to train such a classifier might be different from the one where it is deployed.
\citet{Bakhtin2021} studied this precise setting and showed that the quality of a human/machine discriminator degraded quickly when the prompts used to generate text where from a different domain.
On the other side, the problem of determining the origin of one fixed document is in most use cases not the real problem. 
As an example, an article generated by an existing template-based data-to-text method~\citep{reiter2000} is arguably of less concern than an article diffusing false information written by a human author.
Indeed, one danger of the current language models lies in their capacity of generating a huge amount of human-like text but biased towards a desired opinion or topic~\citep{leroux2020,buchanan2021truth}.

We therefore frame the problem of detecting machine-generated texts as follows: given a  set of documents, and the suspicion that a large fraction of them is generated by machine; is it possible to detect those?
The departure of this setting from the current one is in defining it in a \textit{distributional} and \textit{unsupervised} way.

This setting is complicated by the fact that modern language models coupled with current decoding strategies follow very closely the statistics of generated tokens. 
The key insight of our approach is that, while unigram frequency appears indistinguishable from human generated text, the same is not true for higher-level \textit{n}-grams.
In particular, the distribution of repeats present in the machine-generated text is quite different, as shown in Fig.~\ref{fig:repeatstats}.
While there is apparently nothing special in those repeats (see some examples in Fig.~\ref{fig:repeatsexample}) and they might pass absolutely undetectable when analyzing a single document, they become apparent when looking at a large sample.
Thus, we leverage this signal to pseudo-label the input documents and train an ensemble of classifiers on them. Our experiments show that a majority vote of these classifiers achieves over $90\%$ precision at \numprint{5000} when generating from GPT2-large using \texttt{top-k} sampling, and over $80\%$, for \texttt{nucleus} sampling.

\section{Related Work}
The capacity of large pretrained~\citep{word2vec,BERT} Transformer~\citep{Vaswani2017} models trained only with a cross-entropy loss~\citep{gpt3} are transforming what is possible with natural language processing techniques.
Out of the risks that those models pose~\citep{bender2021} here we are interested in their possibility of generating text which appears as human-generated but without revealing the true authorship.
While those models do commit errors -- ~\citet{dou2021scarecrow} enumerates and analyses 10 types -- they are surprisingly low and often hidden under otherwise very fluent language.
This has created a large line of work that tackles the problem of discriminating such documents, as surveyed in~\citet{jawahar2020}.
In their initial release strategy~\citep{solaiman2019release} OpenAI already mentioned approaches to use the same generation models to identify a text generated with that model, both through fine-tuning on annotated data as well as in a zero-shot setting, using the underlying probabilities.
\citet{ippolito2020} show that an automatic classifier can detect statistical patterns of the generated document and obtains an accuracy for short documents ($60$ tokens) between 70 and 90\% depending on the decoding strategy.
Surprisingly, their performance is different than that of humans, which have a harder time picking up those statistical signals but are unbeaten so far when recognizing incoherence or factually wrong statements.
~\citet{maronikolakis2020} obtains a similar conclusion that automatic models are better than humans for detecting fake headlines.
Studying the performance of automatic classifiers, \citet{Bakhtin2021} provide an exhaustive analysis varying generator and discriminator architecture as well as training data, concluding that performance of the discrimiantor is high if it is trained on data from the same domain.
However, any domain shift affects it in a severe way.
\citet{dugan2020} focuses specifically on the problem of detecting the boundary, from which moment the human generated prompt stops and the machine-written text starts.
The fact that pre-trained language models have statistical patterns which differ from human-generated text has been used in the past not only as feature for detection, but also as input for the interactive tool GLTR~\citep{gltr}.
This tool relies on the probability distribution as given by an inspection model: the case study detailed there shows that even when that inspection model is different than the generation model (although it is of the same family, different sizes of GPT2 to be precise) the highlighted patterns help humans detect the authorship. 
It stops however at unigram features as well.
The concurrent work of \citet{uchendu2021turingbench} proposes a dataset of human and machine-generated samples, across a variety of models, and also frame the problem as authorship attribution.
The proposed baselines are all supervised however, although the authors perform a stylotmetric analysis, concluding the need ``to develop a model that unearth more subtle yet distinct patterns''.

All those studies focus on asking (crowd-sourced) human annotators to decide if a text was generated by a machine or a human.
\citet{clark2021human} points out that the high fluency of modern generation models, combined with a generally low expectation of what machines can accomplish, makes it hard to make this distinguished, even for lightly trained annotators.

\section{Background}

Modern language models are trained through maximum likelihood estimation.
Inferring new sequences is done in an auto-regressive manner, where each token's probability is estimated as a function of its previous context.
Greedy decoding -- this is, picking the most probable token -- is rarely used in open-ended generation.
Similarly, using the probability distribution directly to sample from it is also seldom used: while it mimicks the unigram distribution of tokens very accurately, it often generates incoherent text and makes it therefore easy for humans to detect~\citep{ippolito2020}.
The two most used decoding strategies reduce the universe of possible tokens: \texttt{top-k}~\citep{fan2018} which samples from the $k$ most probable words, and \texttt{nucleus} sampling~\citep{holtzman2019} which samples from the tokens making up the top $p$ probability mass.
Formally, let us assume $c$ is the prefix of fixed windows-size, and which is initially human-written (the prompt) but then gets extended with the generations so far.
$P(\cdot|c)$ is the probability distribution over the vocabulary of tokens $V$ of the language-model, given the context $c$.
These decoding strategies then sample from a re-normalized distribution which modifies the support: instead of being all of $V$, only the most probable tokens are considered. 
This is, we sort all tokens $x^{(1)}, x^{(2)}, \dots, x^{(|V|)} $ according to their probability: $P(x^{(1)}|c) > P(x^{(2)}|c) > \dots > P(x^{(|V|)}|c)$, and define then:

$$ \displaystyle
P^k_\textit{\texttt{topk}}(x^{(i)} | c) = 
\begin{cases}
\frac{P(x^{(i)} | c)} {Z} \text{ if }i \leq k \\
0 \text{ otherwise}\\
\end{cases}
$$
        
$$ \displaystyle
P^p_\textit{\texttt{nucl}}(x^{(i)} | c) =
\begin{cases}
\frac{P(x^{(i)} | c)} {Z} \text{ if } \displaystyle \sum_{j=1}^i P(x^{(j)}) \leq p \\
0 \text{ otherwise}\\
\end{cases}
$$

\noindent where $Z$ denotes the respective normalization constant, and $k$ and $p$ are respective constant which control the sparsity of the support.

\section{Method}
\label{sect:method}

It is well known that current language models trained on large-scale data allow for generation whose individual token probability mimics very closely that of human-generated text.
For example, a frequency-rank plot of text obtained through ancestral or nucleus sampling is almost indistinguishable from that of human text (the famous ``Zipf Law'')~\citep[Figure 7]{nucleus}.

However, the same does not happen for higher-order $n$-grams.
Fig.~\ref{fig:repeatstats} shows a histogram of the length of so-called super-maximal repeats\footnote{A repeated substring $r$ is super-maximal with respect to a collection of documents if there is no other repeat that contains $r$.}~\citep{gusfield1997} for a set of documents generated by each strategy.
Specifically, we consider equivalent sets of documents, where each set is generated using one decoding strategy over a language model, or sampled from human texts (details in Sect.~\ref{sect:experiments}).
The repeats are then computed on each set independently.
As can be seen from Fig.~\ref{fig:repeatstats}, ancestral sampling produces a much larger number of super-maximal repeats.
More interestingly, \texttt{nucleus} sampling not only follows a similar trend of producing more repeats, but that difference gets larger for longer repeats. 
Any individual such repeat can hardly be considered a signal of being machine-generated: in Fig.~\ref{fig:repeatsexample} we depict some examples.
However, from this analysis it appears that they are much more likely to occur in machine-generated text than in a human one.

\begin{figure}
    \centering
    \begin{tabular}{l}
    I could hardly tell what \\
    \hline
    In the latter part of the \\ 
    \hline
    an important figure in the\\
    \hline
    that it is impossible for\\
    \hline
    they had nothing to do with\\
    \hline
    for long periods of time.\\
    \end{tabular}
    \caption{Examples of super-maximal repeats occurring at least 3 times in the set of documents generated with \texttt{nucleus} sampling.}
    \label{fig:repeatsexample}
\end{figure}

\smallskip

Starting from this analysis, we propose a method to detect machine-generated documents in a large collection, based on a self-training approach.
Pseudo-code of this algorithm is shown in Alg.~\ref{alg:detection}.

Given the full collection of documents, we first compute the full set of super-maximal repeats.
A repeated substring $r$ is super-maximal with respect to a collection of documents if $r$ does not occur in any other repeated substring of this collection.
Their number (and their number of occurrences) is linear in (and even bounded by) the combined length of all documents.
They can be computed efficiently in linear time with respect to the total length of the collection using a suffix-tree--like data structure~\citep{gusfield1997}, taking advantage of the fact that $r$ is \textit{not} super-maximal if it can be extended at either side by some symbol and still be a repeated substring: if there is a character $a$ such that $a.r$ or $r.a$ is also repeated, then $r$ is not super-maximal.
In our experiments we used a more memory efficient variation, based on a suffix array~\citep{manber1993suffix}, which in combination of the ``longest common prefix'' (LCP) array allows for fast computation of repeats~\citep{abouelhoda2004replacing}.
Their computation is very fast and only takes a few seconds even for the largest of the collections we consider.

Here we used super-maximal repeats that occur at least $3$ times (to reduce inclusions of spurious repetitions), and have at least 20 characters. 
Note that they are computed on the non-tokenized, raw text and might therefore span several tokens.
Those values were fixed very early on in our experiments, and we did not modify or optimize them further.

A random subset of repeats are then used in order to pseudo-label documents containing those repeats as machine-generated.
The same amount of human-generated documents are then selected.
We explore two scenarios (which impact line~\ref{ln:unsupervised} of Alg.~\ref{alg:detection}): one fully unsupervised, where that selection is noisy, picking from the full collection (and therefore incurring in a $50\%$ mislabeling rate for the negative examples) and one where a separate collection of human text exists.
This pseudo-annotated dataset is then fed as input to a binary classifier.
The set of documents on which this classifier is trained is reduced (maximal $60$ documents, assuming that each occurrence of each of the $20$ repeats occurs in a different document), so we repeat this process $K$ times.
The final raking is obtained by simply summing the predictions of each individual classifier: each document in the dataset is ranked according to the amount of time it was predicted to be machine-generated.


\begin{algorithm}
\caption{Detection of machine-generated documents using repeats.}
\label{alg:detection}
\begin{algorithmic}[1]
\Function{Detection}{document set D, \# of models K}
\State repeats $\gets$ \textit{supermaximal(D, minLen, minOcc)} 
\For{$k \in [1\dots K]$}
\State subset $\gets$ random.choice(\textit{repeats}, 20) 
\State pos $\gets \{ d \in D : \exists r \in \textit{subset} \, :\,  r \in d \} $
\State neg $\gets \textit{random.choice}(D, |pos|) $ // we also experiment using true human text \label{ln:unsupervised}
\State $\textit{clf}_k \gets \textit{train}(pos,neg)$
\EndFor
\State $\displaystyle score_d \gets \sum_{k=1}^K \textit{clf}_k(\textit{machine} | d) >  \textit{clf}_k(\textit{human} | d)  \; \forall d \in D $
\Return $\textit{sort}(\textit{score})$ 
\EndFunction
\end{algorithmic}
\end{algorithm}

\section{Experiments}
\label{sect:experiments}

For the language generator, we decided to use a decoder-only architecture due to its faster inference speed (as opposed to a sequence-to-sequence model like for example T5~\citep{T52020}. 
Because of the large number of generations we needed for these experiments (over \numprint{300000} sequences of of up to $512$ tokens) it was unfeasible to use GPT-3, which at the moment of writing was only accessible through a paid API.
We therefore settled on the largest decoder-only model that could be run on premises: GPT-2, accessed through HuggingFace's Transformers library~\citep{huggingface}.
In order to estimate the evolution of our metrics with increasing model size we used various model.
The main experiments were done with the following decoding algorithms: \texttt{topk} (k=10) and \texttt{nucleus} sampling (p=0.8), and other analysis used greedy decoding and ancestral sampling.

We downloaded the top 100 books from the Gutenberg Project, according to their Frequently Viewed page\footnote{\url{https://www.gutenberg.org/browse/scores/top}} on July 27, 2020, extracted their content and split the texts in paragraphs, keeping only the ones with at least 100 characters.
From there we extracted as prompt the two first sentences of each paragraph.
The continuation was used as human generation (\texttt{textgen}).

The final dataset was composed therefore of continuations of those discarded prompts.
To control for length, we filtered out generations with less than $300$ characters and we trimmed the generated text after the 300th character.
The resulting continuations (between $30k$ and $34k$, depending on the model version) were split evenly to obtain a dataset of machine generated and human continuations.
This is, for one given prompt its possible continuations (either human or machine-generated) appears only once.

We derive from this dataset two different settings:
One, \textbf{semi-supervised}, in which we keep $5\%$ of human-generated documents as negative examples, and another one, fully \textbf{unsupervised}, in which negative samples are chosen at random (and thus, have a 50\% error rate).


Super-maximal repeats were computed on the joint remaining documents, combining both human and machine-generated text.
To produce pseudo-labels of machine-generated text we used super-maximal repeats that are longer than $20$ characters and occur at least $3$ times, and trained classifiers using a distilled BERT model as provided by HuggingFace~\citep{huggingface}.\footnote{class \texttt{DistilBertForSequenceClassification}.} 
We kept those values fixed across all experiments.

Here we report results in 6 settings, corresponding to 3 model sizes (corresponding to GTP-2 released model sizes of - respectively - 124M, 355M and 774M parameters) and 2 sampling strategies (\texttt{topk} and \texttt{nucleus}).

For evaluation, we assume a scenario where our proposed method would be deployed in a semi-automatic system where human annotators would receive a batch of suspicious documents to revise.
This is similar to how current detection of ambiguous and sensitive content is handled (such as fake news or hate speech).
In such a scenario, the accuracy of the final method is of less importance than its capacity to propose a set with high precision of machine generated text.
We therefore take metrics from information retrieval, in particular \textit{precision at $m$} which is the amount of documents that were correctly labeled as machine-generated in the first $m$ documents, when sorting them according to their final score.

\section{Results \& Analysis}
In Fig.~\ref{fig:acck} we report values for \textit{precision at $m$} for the \textbf{semi-supervised} scenario.
The documents are sorted according to the number of times individual experts voted that document to be machine-generated, and we show the results with increasing number of such experts (from 10 to 100).
As can been seen, \texttt{nucleus} sampled text is harder to detect than \texttt{top-k}, something that confirms the literature. 

While there is a clear difference with different sampling techniques, larger models are not always harder to detect than smaller ones, and the final numbers are very close between GPT2-medium and GPT2-large.
Note that in almost all cases, the precision at \numprint{5000} remains higher than $80\%$ using 30 or more experts, and is close to $100\%$ for \texttt{topk}.

Finally, Fig.~\ref{fig:acckNoise} reports the same results when using noisy data not only for machine generated but also for human-generated documents.
This follows the \textbf{unsupervised} scenario where no other in-domain human generated text is available.
While the numbers are significantly lower, the precision at $m$ remains high for \texttt{topk}, and is around $80\%$ for \texttt{nucleus} sampling on GPT2-large for the first $100$ documents.

In Fig.~\ref{fig:examples} and~\ref{fig:examplesWrong} we show -- respectively -- correct and incorrect examples of the top-25 documents obtained through \texttt{nucleus} sampling on GPT2-Large, using noisy labels for the human-generated text (the case of Fig.~\ref{fig:acckNoise}).
Fig.~\ref{fig:examples} are indeed machine-generated, while Fig.~\ref{fig:examplesWrong} are human-generated and therefore wrongly retrieved.
Given an individual example, there does not seem to be nothing apparent which would allow to judge on their origin.
It is the distributional approach that obtains a signal by looking at a large collection of such documents.

\begin{figure}
    \centering
    \input{examples}
    \caption{Examples of sentences that were classified \textbf{\bgg{correctly}} as machine-generated, from the top-25 generations with \texttt{nucleus} sampling using GPT2-Large.}
    \label{fig:examples}
\end{figure}

\begin{figure}
    \centering
    \input{examples-wrong}
    \caption{Examples of sentences that were classified \textbf{\bgr{incorrectly}} as machine-generated, from the top-25 generations with \texttt{nucleus} sampling using GPT2-Large (this is, these are human-generated).}
    \label{fig:examplesWrong}
\end{figure}

\begin{figure*}
    \centering
    \includegraphics[width=\textwidth]{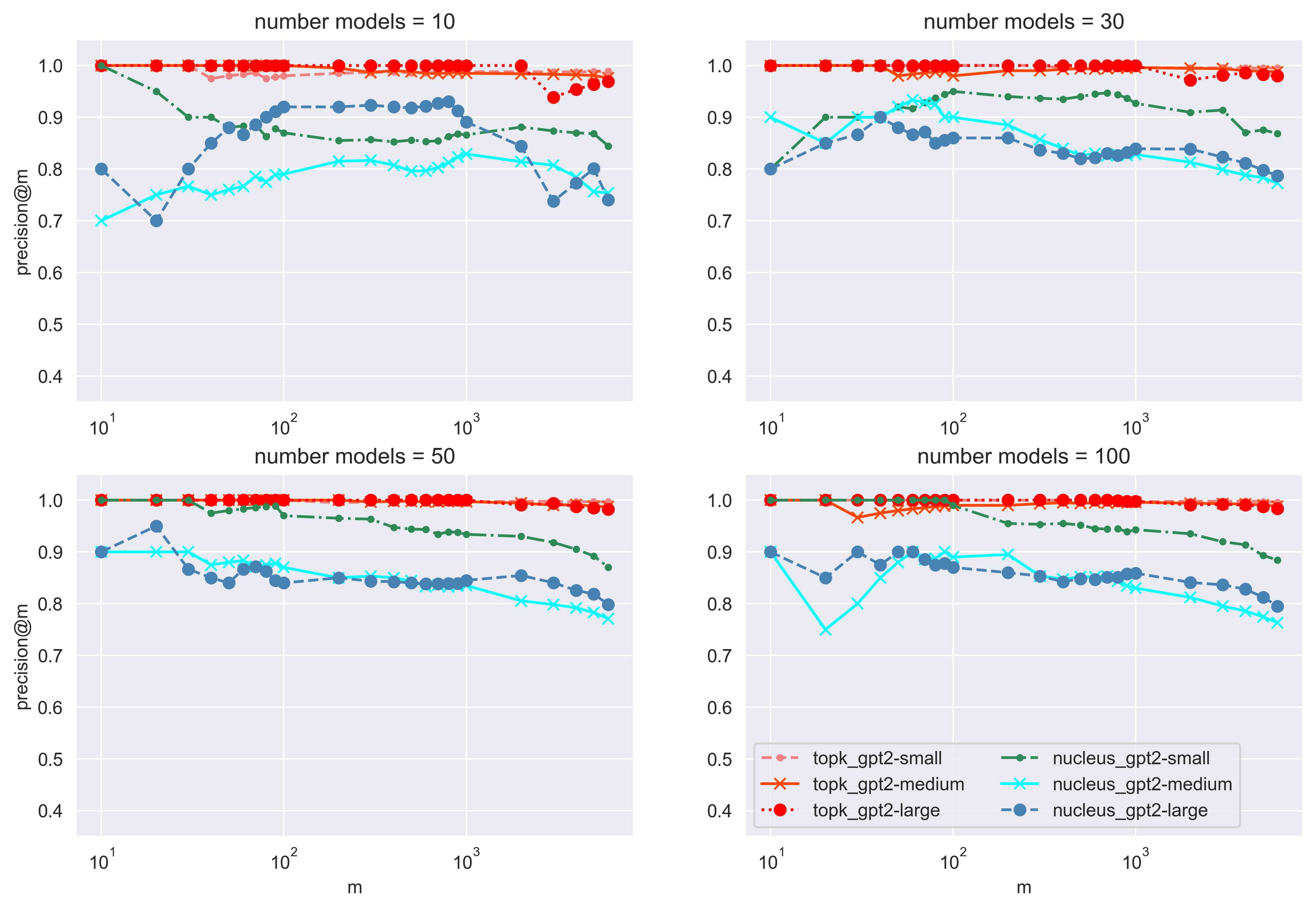}
    \caption{Precision at $m$ for various models and two sampling strategies on the semi-supervised scenario.}
    \label{fig:acck}
\end{figure*}

\begin{figure*}
    \centering
    \includegraphics[width=\textwidth]{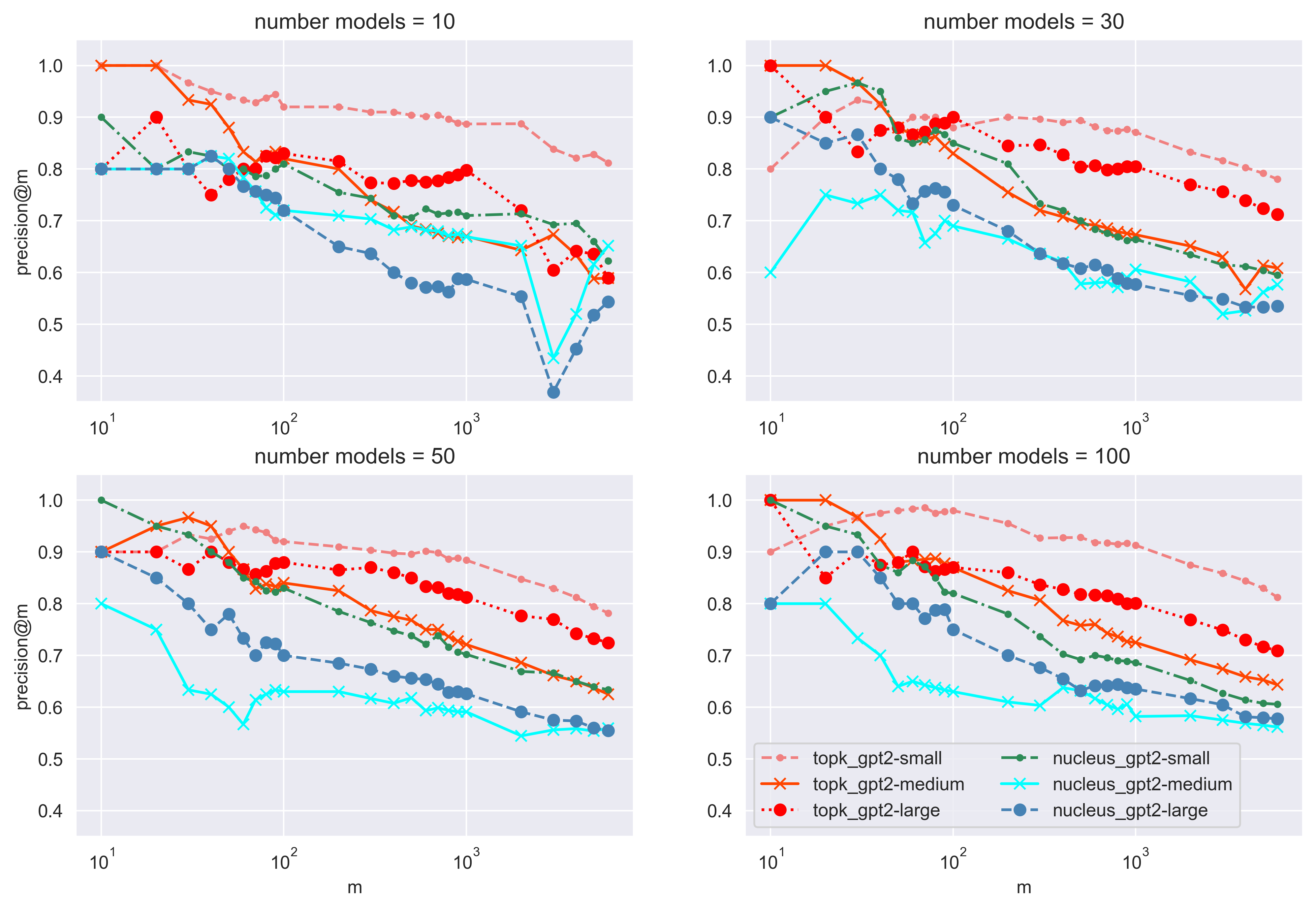}
    \caption{Precision at $m$ for various models and two sampling strategies in the fully unsupervised scenario.}
    \label{fig:acckNoise}
\end{figure*}

\subsection{Ablating pseudo-relevance loop}
The use of repeats to capture a weak signal raises the question if the pseudo-relevance step is necessary, or if classifying all those documents that contain a supermaximal repeat is not a good enough baseline.
A first issue is that such a setting does not allow for fine-grained scoring: in all the datasets no document contained more than one supermaximal repeat.
Precision at $m$ is therefore not applicable, as the sorting only throws documents into two bins, depending if they contain or not such a repeat. 

In Table~\ref{tb:ablation} we report the percentage of machine-generated documents out of all those that contain supermaximal repeats, for the various combination of decoding algorithm and model size. 
In all cases those numbers are lower than what can be obtained after training a classifier on those documents (Fig.~\ref{fig:acck} and~\ref{fig:acckNoise}) pointing towards a better discrimination power by going through an additional classifier step.
For example, for nucleus sampling the percentage of documents containing a supermaximal repeat that are machine-generated is around $60\%$, while it is over $80\%$ with the additional classification step in the weakly supervised case (and around $70\%$ for comparable values of $m$ in the unsupervised one).
A similar absolute difference -- for higher values -- can be observed for top-$k$ sampling.

\begin{table}
\centering
\begin{tabular}{|cccc|}
\hline
decoding & model & prec@$m$ & $m$ \\
\hline
\multirow{3}{*}{nucleus} & small & $60.3$ & \numprint{5745} \\
 & medium & $58.0$ & \numprint{4705} \\
 & large & $57.6$ & \numprint{4856} \\
\hline
\multirow{3}{*}{top-$k$} & small & $75.0$ & \numprint{9095} \\
 & medium & $71.7$ & \numprint{7049} \\
 & large & $71.3$ & \numprint{7208} \\
\hline
\end{tabular}
\caption{Precision at $m$ values by classifying all documents containing super-maximal repeats as machine-generated.}
\label{tb:ablation}
\end{table}

\subsection{Full Classification}

\begin{figure}[bh!]
    \centering
    \includegraphics[width=.5\textwidth]{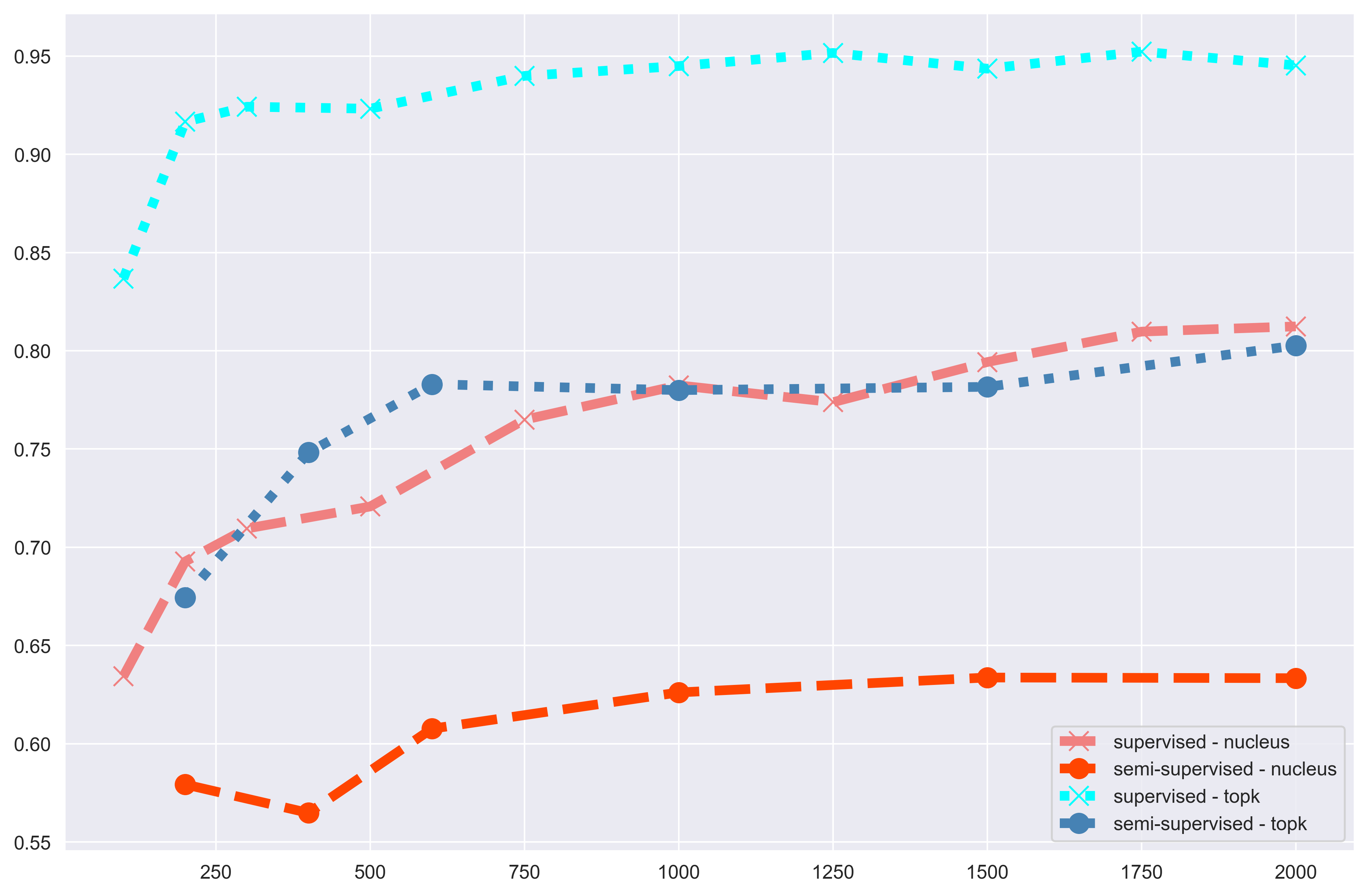}
    \caption{Self-supervised (only using human-generated texts) approach for a classifier, compared to a fully supervised approach.
    The semi-supervised approach used $K=10$, and in all cases the text was generated by GPT2-Medium and the binary classifier is a fine-tuned distilled BERT model.
    }
    \label{fig:supervised}
\end{figure}

The approach described so far permits a semi-automatic human-in-the-loop system, where the classifier selects highly probable documents for further human inspection.
However, the self-training approach can be pushed even further, to train a classification model that aims to classify the whole data-set.
A standard approach in information retrieval to enrich the queries is \textit{pseudo-relevance feedback}: given a query, this approach assumes that the top-most documents are relevant.
A second query is then created, which takes the original one and enriches the query with terms from those top-ranked documents.
The intuition is that those alternative terms allow for richer information and might capture documents that would otherwise be missed.

Inspired by this, we aim to train a classifier that uses the top ranked documents of Alg.~\ref{alg:detection} and measure its accuracy on a fixed held-out data-set.
In Fig.~\ref{fig:supervised} we plot the resulting accuracies with increasing number of training data.
To create this pseudo training data we use again gold human-generated documents, while documents labelled as machine generated are the top ranked documents by the semi-supervised approach of Sect.~\ref{sect:method}.
The dataset here is obtained with GPT2-medium.
As comparison, we also report the results obtained with a fully supervised model on the same test data and the same number of training data.
The gap between the fully supervised and semi-supervised approaches is high for both decoding strategies.
However, the semi-supervised system only had gold labels for human-generated text and all the machine-generated labels were inferred from the data.

\section{Discussion}

Why does this work at all?
Existing zero-shot detection approaches, like the analysis done with 1.5B-GPT2 of \citet{solaiman2019release} or the GLTR interactive tool~\citep{gltr}, rely on the probabilities obtained by an inspection models, which in most cases have some familiarity with the generation model.
The close distribution of unigram frequencies of generated text might indicate that modern language models do not seem to have a clear statistical signature, a reason why the text generated by them is often hard to detect with classifiers. 

However, the results in this work seem to indicate that those patterns are still present, but at the level of repetitions.
Assuming that the language models is one author, this work can be seen as detecting the single author who generated $50\%$ of the dataset.
The fact that this is (somehow) possible could be an indication that the prompt is not sufficient for the language-model to mimic the original author's style, and that some of its own style -- evidenced by the repetition of some phrases -- becomes apparent.
Further studies in the stylometry of the generated text and how this relates to the length and type of the prompt might provide more insights into the limitations and adaptability of large-scale language models.

\begin{figure*}[th!]
    \centering
    \includegraphics[width=300pt]{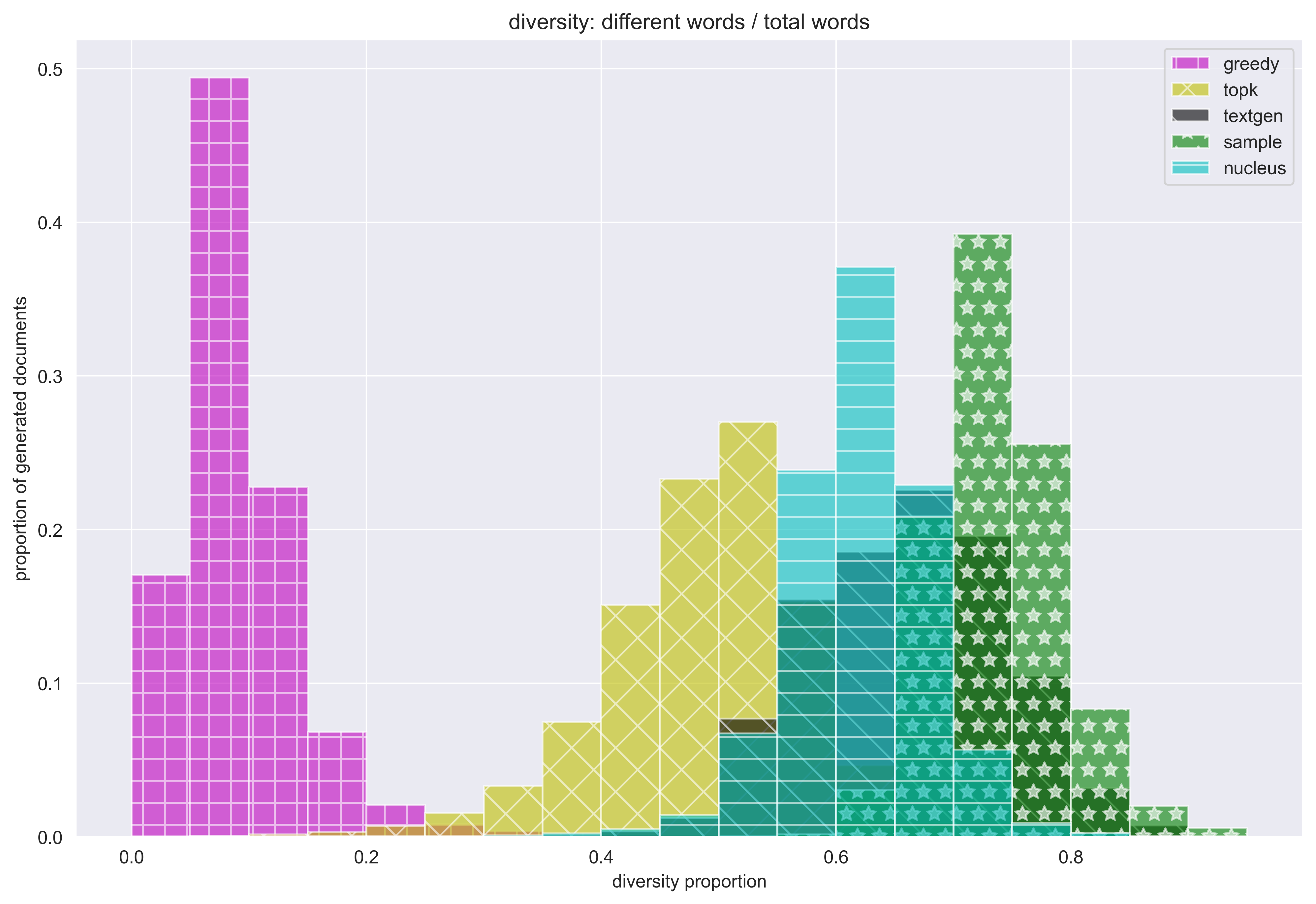}
    \caption{Histogram of diversity for different sampling strategies.}
    \label{fig:diversity}
\end{figure*}

Another related cause might be due to the \textit{diversity} of the generated text.
Fig.~\ref{fig:diversity} 
plots the relative histogram of the diversity, measured as the number of different words divided by the number of tokens in the document.
As is well-known, \texttt{greedy} produces unvarying continuations and is prone to repetitions. 
The other sampling techniques have a much closer distribution to human text (\texttt{textgen}).
Human text, however, is less peaked than the sampling strategies, including ancestral sampling (\texttt{sample}) which -- in average -- produces even more diverse text.
That diversity has a direct impact on the number of repetitions: a collection of documents with lower diversity is more probable to give rise to longer repetitions. 
This might be another reason why \texttt{top-k} and \texttt{nucleus} have a larger set of such long repetitions.
Moreover, note that the detection accuracy is higher for \texttt{top-k} than for \texttt{nucleus}, something that correlates with their diversity.

\section{Conclusion \& Limitations}
We analyzed characteristics of machine-generated text, detecting that they tend to generate certain (well-formed) phrases more often than in human text.
Using that signal as an input to a series of classifier our experiments shows that this signal can be amplified to allow to rank documents in such a way that machine generated documents tend to appear in the beginning.
A limitation of this work is the balanced assumption of the training set, as well as the somehow rigid nature of exact repeats (which does allow fast computation however).
We believe that further stylometric analysis of generated text of such models could allow to provide additional warning signals that machine-generated content is flooding a certain corpus and even identify examples of such documents.

\bibliography{biblio}
\bibliographystyle{acl_natbib}




\end{document}